# AI-Guided Defect Detection Techniques to Model Single Crystal Diamond Growth

Rohan Reddy Mekala[1,a], Elias Garratt[2,b], Matthias Muehle[3,c], Arjun Srinivasan[1,d], Adam Porter[4,e] and Mikael Lindvall[1,f]

[1]Fraunhofer USA, Center Mid-Atlantic, US

[2]Michigan State University, US

[3]Fraunhofer USA, Center Midwest, US

[4]Fraunhofer USA, Center Mid-Atlantic & Department of Computer Science, University of Maryland, US

[a]rreddy@fraunhofer.org, [b]garrate@msu.edu,
[c]mmuehle@fraunhofer.org, [d]asrinivasan@fraunhofer.org, [e]aporter@fraunhofer.org,
[f]mikael@fraunhofer.org

**Keywords:** Diamond growth, AI in Material Science, Deep learning, Data Labeling, Active learning

**Abstract.** From a process development perspective, diamond growth via chemical vapor deposition has made significant strides. However, challenges persist in achieving high quality and large-area material production. These difficulties include controlling conditions to maintain uniform growth rates for the entire growth surface. As growth progresses, various factors or defect states emerge, altering the uniform conditions. These changes affect the growth rate and result in the formation of crystalline defects at the microscale. However, there is a distinct lack of methods to identify these defect states and their geometry using images taken during the growth process. This paper details seminal work on defect segmentation pipeline using in-situ optical images to identify features that indicate defective states that are visible at the macroscale. Using a semantic segmentation approach as applied in our previous work, these defect states and corresponding derivative features are isolated and classified by their pixel masks. Using an annotation focused human-in-the-loop software architecture to produce training datasets, with modules for selective data labeling using active learning, data augmentations, and model-assisted labeling, our approach achieves effective annotation accuracy and drastically reduces the time and cost of labeling by orders of magnitude. On the model development front, we found that deep learning-based algorithms are the most efficient. They can accurately learn complex representations from feature-rich datasets. Our best-performing model, based on the YOLOV3 and DeeplabV3plus architectures, achieved excellent accuracy for specific features of interest. Specifically, it reached 93.35% accuracy for center defects, 92.83% for polycrystalline defects, and 91.98% for edge defects.

## Introduction

The ability to create large single crystal diamonds (SCD) at high quality (low dislocation density, high chemical purity, and smooth surfaces) is a longstanding and very high-value challenge for the scientific community, and one that has yet to be surmounted despite over decades of sustained materials development effort. Moreover, the ability to produce diamonds at scale poses a significant future challenge for manufacturers of electronic-grade or quantum-grade diamonds. This challenge stems from the absence of existing techniques to predict future states or features of the diamond during the growth process, which hinders the implementation of prescriptive measures for process control or correction. While diamond is highly competitive in power electronics, quantum, and health sciences and engineering fields, overcoming these challenges in quality and size is essential for its continued competitiveness. Commercial adaptation of existing manufacturing infrastructure and pipelines will require reliable access to high-quality SCD wafers surpassing 1 to 2 inches in lateral size to be compatible with existing semiconductor fab lines. Current paradigms in diamond material process development schemes rely on human operators to design new experiments through trial and error and lead to large stochastic results in product yield, along with often unexplainable macro and micro defects. Therefore, new approaches to accelerating the material process development cycle are needed. Of these, machine learning presents a compelling solution in predicting future states of diamond growth.

Computer vision-driven AI algorithms, in general, excel in feature extraction tasks such as image segmentation, object detection, and spatiotemporal feature prediction. With advancements in deep learning, these techniques can generalize over large datasets, capturing complex patterns between input and output data. In this paper, we introduce the first utilization of machine learning and deep learning (ML/DL) algorithms [1] to extract and analyze spatial defect features from in-situ image data collected during independent diamond growth runs and demonstrate that ML/DL-based models offer an effective solutions for prediction objectives in safety- critical systems [2], making them ideal for diamond process development. Complementing our work on geometric macro-feature extraction, these algorithms enhance our ability to predict features in future states of diamond growth, ultimately shortening the material development cycle and improving wafer sizes and crystalline quality.

**Contributions**

This paper outlines research and development of a novel pipeline based on DL-driven semantic segmentation and object detection algorithms to extract independent and derivative defect features of interest from time sequenced images taken during diamond growth. Through development of the proposed DL-backed pipeline, we advance the state-of-the-art defect detection techniques in diamond growth in the following ways.
1. This is a novel attempt at the successful development of a DL-based pipeline for defect detection in the diamond growth domain for low-volume high-feature- complexity training dataset environments where both data procurement and data annotation are extremely expensive tasks. That is, our approach is designed with the time and expense of growing diamond in CVD reactors and complex undesirable defects that occur during growth, like edge defects, center defects and polycrystalline growth. Additionally, our approach achieves excellent pixel-level and bounding-box-based prediction accuracy metrics and serves as the first benchmark established for automated defect extraction in the domain of diamond growth.
2. This is a novel crowd-backed labeling pipeline designed, implemented, and validated for creating defect-annotated image datasets for semantic segmentation and object detection in diamond growth. Our implementation significantly reduces labeling time and costs (from 15 minutes to 2 minutes per training image), ensuring consistency, accuracy, and integrity of labeled data.
3. Moreover, our DL-based pipeline, using our best performing model, achieved excellent defect detection accuracy level Average Precision (AP) [3] of 93.35%, 92.83% for the Center Defects and Edge Defects respectively, Intersection over Union (IoU) accuracy [4] of 91.98% for the polycrystalline defects.
4. This is a novel attempt at determining and comparing accuracy measures for diamond growth defect detection objectives against a near-exhaustive range of state-of-the-art DL models.
5. This paper extends previous objectives by evaluating model permutations based on variations in input image resolution and dataset size. It showcases insights gained from experiments to enhance output precision of defect detection models, demonstrating the potential for improved accuracy metrics with increased image resolution and training data volume for specific model architectures.

Our research is the first step towards deep learning as a defect detection mechanism in guiding the automated analysis of diamond synthesis pipelines. The benefit of high detection accuracy, and extremely low compute times, without relying on voluminous labeled data for the training sets should serve as a foundation to further develop the design and methods proposed in the paper for better defect detection benchmarks in the diamond growth domain.

**Related Work**

The field of defect detection, while novel to the domain of diamond growth in general, has been previously tackled for sub-domains within manufacturing like metals processing and additive manufacturing. These implementations used to solve defect detection objectives have traditionally been employed across the categories described below.

**Computer Vision and Machine Learning based Methods in Defect Detection.** In manufacturing, both computer vision and machine learning play crucial roles in defect detection. Traditional computer vision techniques, such as image processing and morphology, have been effective in identifying defects on surfaces like rolled steel and ceramics [5,6]. However, they often require extensive manual feature engineering and struggle with complex defects. In contrast, machine learning methods offer promising alternatives. Support Vector Machines (SVM) and K-Nearest Neighbor (KNN) algorithms [7] have successfully detected defects on iron castings [8], while shape-matching algorithms combined with machine learning have shown potential in discovering new crystal structures [9]. Despite their successes, both traditional computer vision and machine learning methods face limitations with highly complex defects, often falling short in tasks requiring precise object detection and segmentation [10]. These challenges underscore the ongoing need for innovative approaches to defect detection in manufacturing.

**Deep Learning Methods in Defect Detection.** DL-based methods, renowned for their exceptional performance in state-of-the-art accuracy metrics, have been employed in the past for applications involving automated inspection and quality monitoring. For instance, DL has been used to segment defects in solar cell manufacturing [11] and to classify defects in metal components produced from additive manufacturing [12] with excellent results. DL has also been used in the field of crystallography for classification and prediction of crystal structure [13]. However, a notable challenge with DL approaches lies in their dependence on large, labeled datasets, which can be costly and time-consuming to procure and annotate. In this research, we address these challenges by proposing a DL-based pipeline for data selection, annotation, and model development tailored for low-volume training dataset environments. Our approach achieves superior semantic segmentation and object detection accuracy while mitigating the resource-intensive nature of traditional DL methods.

## Background

The conditions of the growth surface of the diamond growth surface change over the course of growth due to several factors. These include getting closer to the plasma, the
growth of parasitic polycrystalline diamond material, and expansion of the growth surface. These can result in the formation of defects in diamond, which degrade its quality and can make it unsuitable for technological applications.

Therefore, AI-models of diamond growth must incorporate information on the quality of the growing diamond crystal. This kind of data can be obtained in several ways, including through image capture and analysis. Image capture is an attractive option because it is an inexpensive, fast, data rich method of providing proof-of-concept defect identification and analysis. In this study, defects are classified based on their shape, shape features, and color using high resolution digital images collected with an 18-megapixel DSLR camera. Each combination of these are characteristics of problems at the growth surface which can give rise to, or indicate the presence of, crystal defects.

In subsequent sections we will detail the development of object detection and segmentation models which classify and isolate accurate pixel masks of defects. We will provide details on defect detection pipelines developed using techniques ranging from traditional computer vision techniques to advanced deep learning-based methods and establish benchmarks on accuracy metrics achieved on a general and defect label-wise level.

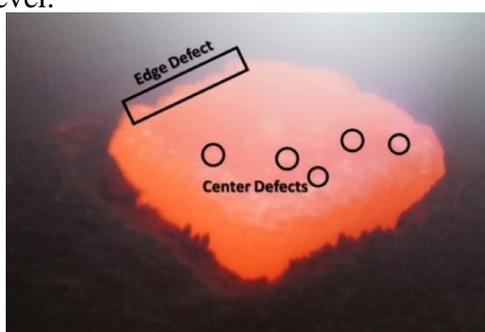
Figure 1. Independent and Derived DOIs for the detection pipeline

**Defects of Interest in Modelling Diamond Growth.** On a broad level, our defect detection pipeline focuses on extracting the following defect label categories illustrated in Figure 1. Defects in general can be classified into microscale and macroscale defects, however, macroscale defects is the focus of this work due to limitations of the reactor and cameras used to collect data. As a rule, macroscale defects are indicative of underlying problems at the microscale, like lattice mismatching which can form grain boundaries, and edge and slip dislocations. By training AI models to recognize macroscale defects such as the overall shape or smoothness of growth edges, they can accurately assess quality and classify growth quality. This is expected to enhance production quality in the short term and enable prediction of quality in the long term. These defect classification categories enable AI models to recognize defective conditions of growth. That is, any condition where the result leads to poor crystalline quality. In the short-term, models are not intended to diagnose the relative goodness or badness of the defect. However, in the long-term, models will be able to use aggregated data on macroscale defects to rate growth processes and predict growth outcomes. In this work, macroscale defects subsets are broken down into three categories, or DOIs, with each contributing to non- uniform conditions near or at the growth surface that may be detrimental to single crystal diamond growth.

1. Polycrystalline Defects - Category 1 defects of interest (DOIs) are polycrystalline diamond growth, either on the holder or on the diamond. Since we are interested in areal density of spread in this class of defects, we modeled detecting them as a semantic segmentation objective. Growth on the holder is common and contributes to non-uniform growth at the diamond substrate surface in two ways. The first is robbing the substrate growth surface of gas radicals which grow diamonds, which negatively affects the growth rate. The second is creating field effects and temperature variations near the substrate growth surface, which also affects the growth rate. Relative to the substrate growth surface at the center, this leads to unpredictable variation in growth rate with results that include shrinkage of single crystal surface over time and formation of non-polyhedral macroscale shapes. Non-polyhedral shapes contain numerous defects. These are caused either through polycrystalline diamond formation directly forming on the diamond and replacing single crystal surface areas, or by polycrystalline diamond forming on the holder and growing into the pocket holder, which in turn constrains the single crystal diamond growth.

2. Center and Edge Defects - Category 2 DOIs are highly localized spots on the growth surface of the diamond substrate and rough edges, or edges on a geometric shape that do not form smooth linear boundaries. Spots appear as white dots against the orange-red substrate background and from blackbody radiation, white spots are higher temperature areas where temperature relative to the surrounding surface is much higher. Such localized temperature variations can result in localized variations in growth rates and lead to small concentrations of defects. Non-smooth boundaries, like those between the orange-red diamond substrate and background (Figure 1) also indicate that growth conditions are not uniform, though this may be due to other factors in addition to temperature like an initial rough surface. Since we are interested in the number of both center and edge defects present it is modeled as an object detection problem.

3. Shape Category 3 DOIs are overall diamond shape. A uniform balance of temperature and available gas radicals will typically lead to polyhedral crystal growth, so long as the system is not constrained by outside forces - i.e., circular silicon wafers constrain growth to the circular shape of the silicon wafer at the macroscale. Therefore, quality crystal growth can be classified by object type, for diamond this is the formation of octohedral shapes over long growth time periods If growth shape assumes a non-octohedral or square top surface in the pocket holders we use, this can be classified as poor growth. It should be noted that such results can be due to holder design, polycrystalline growth/growth rate, plasma conditions (ratio of gas radical species), plasma temperature, and more.

**Image Recognition Task of Interest.** For the purposes of this paper, three classes of techniques were considered for the task of DOI extraction, 1) Object Detection, comprising a broad group of algorithms that classify patches or pixel-groups of an image into different object classes and predict a bounding box corresponding to that object, and 2) Semantic Segmentation, comprising a broad group of algorithms that classify each pixel within an image as multiple classes with the constraint of not being able to differentiate between multiple instances of the same class, and3) Instance Segmentation, comprising a broad group of algorithms that classify each pixel within an image as multiple classes, where the model can also tell apart instances of the same class.

For defects like polycrystalline growth, since knowing pixel-level changes in crystalline defect properties on the surface is crucial towards understanding the growth characteristics of the diamond over subsequent frame prediction models, it is imperative for the defect detection model to be able to classify the presence of these defects within the input image on a pixel-wise level with high precision and accuracy. This renders object detection algorithms ineffective for polycrystalline defects as they predict rectangular bounding boxes focused on detecting the "number" of objects instead of the pixel-wise area. Additionally, as diamond growth is a relatively slow process, we observe a limited increase in independent DOI pixels over time, making it even more imperative to develop models that can deliver pixel-level classification capabilities for this class of defects. For the above-mentioned reasons, we chose pixel-level semantic segmentation as the image recognition objective of choice for use in the research and development of our model architectures for polycrystalline defects.

For edge and center defects on the other hand, since accurately ascertaining change in "number" of defects over time was more important than measuring areal change, we settled for object detection as the image recognition objective of choice. Object detection algorithms are designed to minimize variance between predicted and actual levels while being more computationally efficient than segmentation algorithms. This makes them ideal for recognizing edge and center defects in images. In the following sections, we explore different algorithms tested for developing models for both object detection and semantic segmentation.

1. Traditional computer vision methods for semantic segmentation rely on color/region-based thresholding and edge detection. Object detection use algorithms based on edge templates, nearest neighbor concepts, Haar wavelets, and SVM techniques [14]. While enhancements like feature descriptors such as SIFT and multi-scale detection cascades have improved accuracy and speed, these methods still heavily rely on hand-crafted features and manual tuning. Consequently, they are inadequate for detecting complex features in diamond growth images.

2. Statistical machine-learning methods for segmentation, such as k-Means clustering and superpixel segmentation, also require human intervention for feature clustering and fine-tuning. Due to their inability to handle complex feature distributions and extreme spatial variations, particularly in diamond growth images, these traditional methods prove ineffective. Neural networks and deep learning, with their nonlinear learning capabilities, offer superior performance in capturing and generalizing complex patterns in images, which we will explore further in subsequent sections [15].

3. Deep learning techniques, particularly convolutional neural networks (CNNs), have been instrumental in our research on defect extraction within the diamond growth domain. We have focused on semantic segmentation, leveraging architectures like Fully Convolutional Networks (FCN) [16], DeeplabV3 [17], and DeeplabV3Plus [18]. In our experimentation, DeeplabV3Plus emerged as the top performer, incorporating separate decoder modules and dilated depthwise separable convolutions for enhanced feature extraction. This architecture employs Xception [19] as its backbone, reducing computations while maintaining accuracy. For object detection, our initial models utilized dual-stage approaches such as RCNN [20] and FasterRCNN [21], with subsequent exploration into single-stage detectors like YOLOV1 [22] and its iterations. Among these, YOLOV3 [23] stood out for its deep architecture and skip connections, proving superior for defect label detection in the diamond domain.

The advancement of deep neural networks, especially CNNs, has advanced defect extraction in the diamond growth domain. Our focus has been on semantic segmentation, employing advanced architectures like DeeplabV3Plus for precise feature extraction. Additionally, for object detection, we explored both dual-stage methods like Faster- RCNN and single-stage detectors like YOLOV3. Notably, YOLOV3's deep architecture and skip connections have shown exceptional performance in defect label detection tasks, making it the preferred choice for our experiments.

**Method And Design of Experiments**

We devised an innovative AI-driven pipeline for defect extraction from time-sequenced reactor images in diamond growth processes, aiming to provide post-analytics on the growth process. Our pipeline leverages deep-learning architectures to achieve state-of-the-art segmentation and object detection accuracy metrics within the constraints of limited training data. The models developed require annotated images to train, which poses challenges due to the complexity of defects and the need for diverse training data representative of real-world scenarios. Obtaining such data and annotations is both costly and time-consuming. Therefore, our pipeline includes intelligent data selection, augmentation, and labeling components, with feedback loops involving domain experts, labelers, and models under training. The following sections detail our pipeline, covering modules for data collection, pre-processing, model research and development, evaluation, and post-analytics.

**Growth Run Data Procurement Experiment Module, SP1.** As part of the data procurement module, we collected images for 25 growth runs in the RAW format using a full-frame mirrorless interchangeable lens camera (Sony Alpha 7R IV), equipped with a macro lens (Sony SEL90M28G) via an optical viewport of the reactor's electromagnetic cavity. On average, data from each growth run procured spans 34 hours of image and reactor telemetry data (i.e., substrate temperature, reactor pressure, gas flow rates). Images were collected at a 1-minute frequency, while reactor telemetry data was logged every second. Data was synced to a remote storage server for coordination between the manufacturing and data science teams. Reactor configuration and conditions were kept within previously mapped, stable growth regimes [24].

**Data Pre-processing Experiment Module, SP2.** The data pre-processing module prepares image datasets for annotation and machine learning model development in two stages. Initially, time-sequenced image data is resampled into 15-minute windows and converted to PNG format to minimize information loss. Traditional computer vision techniques are then employed to filter out noisy or blacked out images. Subsequently, 300 images are selected for further processing by the labeling module, SP3. Despite the relatively small dataset size, we achieved state-of-the-art accuracy by implementing a robust framework with custom sub-modules for active-learning-based data selection, augmentation, annotations, and model development.

In the second stage, the datasets are fine-tuned for machine learning model development by cropping images to bounding boxes containing individual DOI contours and resizing them to 256x256 and 512x512 resolutions. Further preprocessing involves denoising and normalization. The dataset is then randomized and split into training and test sets in a 90:10 ratio. The training set, once annotated, is used for model development and training, while the test set evaluates model performance. Although a larger dataset would allow for more experimentation with cross-validation techniques, our 90:10 split maximized training data size and yielded slightly better results compared to an 80:20 split.

**Data Labeling Experiment Module, SP3.** The AI-based human in the loop data labeling module to develop annotations for DOIs within the image datasets is a critical component of our overall pipeline and pivotal in solving for the issues concerning low volume datasets and the expensive nature of procuring both the images and developing the corresponding annotations in a fast and accurate manner. Using the module developed, we were able to reduce the average labeling time for each image from 13 minutes to 3.2 minutes.

The module sub-pipeline, as portrayed in Figure 2, starts with saving the pre-processed image data from the SP1/SP2 modules as the input image database for use in annotation development. On average, when using the labeling platform in isolation, labeling each image accurately for the geometry-based semantic segmentation and object detection objectives takes 15 minutes. Given this time-intensive nature of annotating images, we customize selection of images using a selection algorithm based on active-learning [25] to only label a subset of the images that potentially increase the segmentation model accuracy metrics towards corresponding DOIs. The selected images are then passed on in batches of 100 images each to the labeling process.

For the consequent labeling process, we use a crowdsourced labeling platform for image recognition tasks called LabelBox [26] to label the growth-run images. The initial batch of 100 images is reviewed by a team consisting of 3 material scientists and 15 external labelers sourced from Labelbox's crowdsourced labeling team. Material scientists provide detailed instructions through explanatory videos and meetings, facilitating a collaborative annotation process. Annotations undergo crowd-sourced peer review, where predominant occurrences are identified based on a consensus score. Subsequently, material scientists conduct a final review to ensure accuracy, consistency, and domain-level data integrity.

To address variability in annotation instructions and interpretations among labelers, we implement model-assisted labeling (MAL). A baseline model is trained incrementally over iterative image-annotation pairs, with updated model outputs overlaid on subsequent batches for annotation. This approach significantly reduces labeling time while enhancing label consistency and accuracy.

Additionally, the time to label for each image reduces significantly from 15 minutes to 2 minutes. This loop of model assisted data annotation and labeler-guided correction [27] is also continued until a threshold segmentation accuracy of 80% is reached for the model being used towards MAL. Once the threshold accuracy is reached, all future batches of images selected for annotation use contour overlays from the finalized baseline model as a starting point for annotation by the labelers.

Once the final set of images-label pairs meet the requisite requirements on the minimum number of images processed, they are passed over for use in research and development of the final version of the semantic segmentation and object detection models, as explained in the next module.

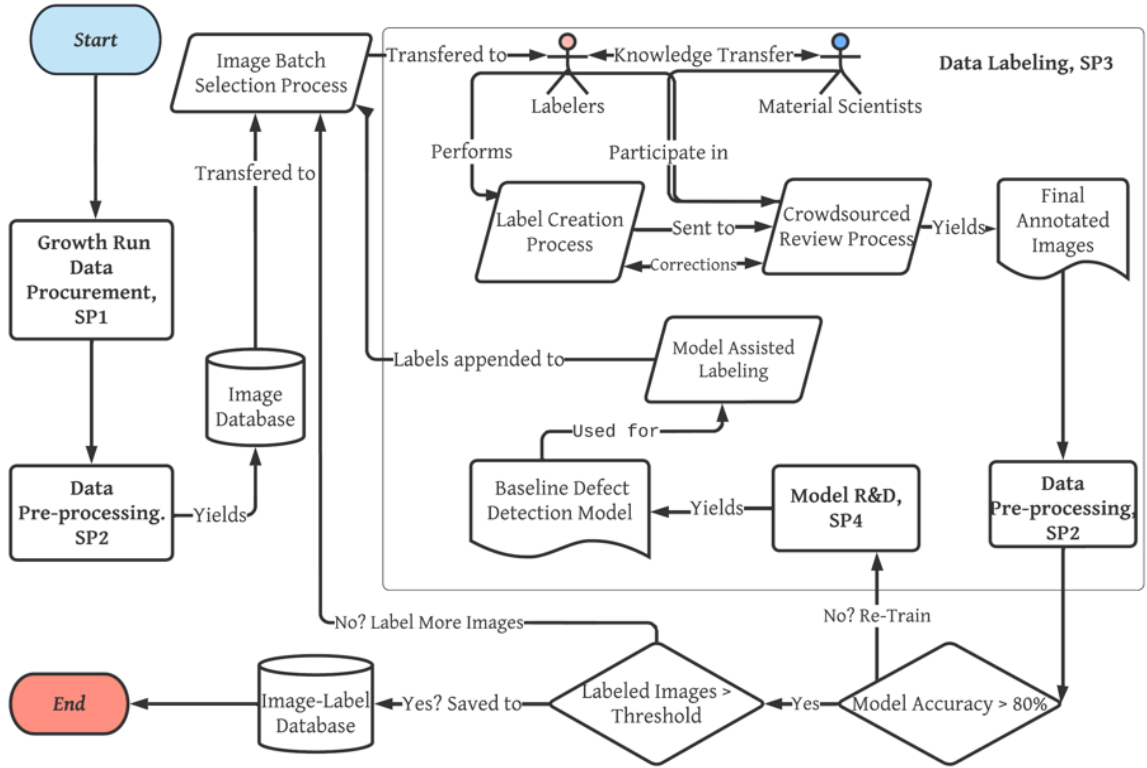

Figure 2. Flow chart explaining the Data Development and Labeling Modules

**Model Research and Development Experiments, SP4.** The model research and development module comprise the use of image-label datasets from SP3 for development, optimization, testing, and benchmarking of various DL-based model architectures, in pursuit of finding the optimal parameter set for DOI detection and classification for the domain of diamond growth. To develop and test different DL-based model architectures mentioned, we used tensorflow [28], a deep learning library with strong visualization capabilities and dynamic, near-exhaustive options for use in deep learning and computer vision-driven model development. The segmentation models were trained on a machine with 32GB RAM, a 12-core 3.50 GHz processor, and an NVidia GeForce RTX 2080 Ti graphics card with 12GB VRAM.

Our research focuses on developing model architectures for object detection and semantic segmentation using image- label datasets obtained from the SP3 module. YOLOv3- based architectures are selected for object detection, while Deeplabv3Plus is chosen for defect segmentation. We employ Darknet-53 and a modified version of Xception as backbone architectures for YOLOv3 and Deeplabv3Plus models, respectively. Training involves 30–45 epochs with a batch size of 20 and learning rates between $6 \times 10^{-6}$ and $3 \times 10^{-4}$. For loss optimization, we experiment with various loss functions, including sparse categorical cross entropy and focal loss[29]to effectively handle class imbalances.

In addition, we explore evaluation metrics such as pixel accuracy, mean intersection over union (mIoU) [4], and average Precision (AP) [3] for benchmarking segmentation and object detection models. mIoU is chosen as the primary evaluation metric due to its ability to provide unbiased accuracy representation for datasets with class imbalances, while AP is used for object detection models to measure detection precision under different recalls. The module development is carried out in two broad steps. The first step is to obtain a baseline model with good generalization performance, iterating over a subset of the image datasets till a benchmark classification accuracy of 80% is achieved. This model is iteratively improved as part of the model-assisted labeling process explained in the previous section. The second step comprises further refining and development of the model for the remaining batches of the image-label datasets created, to set benchmark DOI classification accuracies for the diamond growth domain.

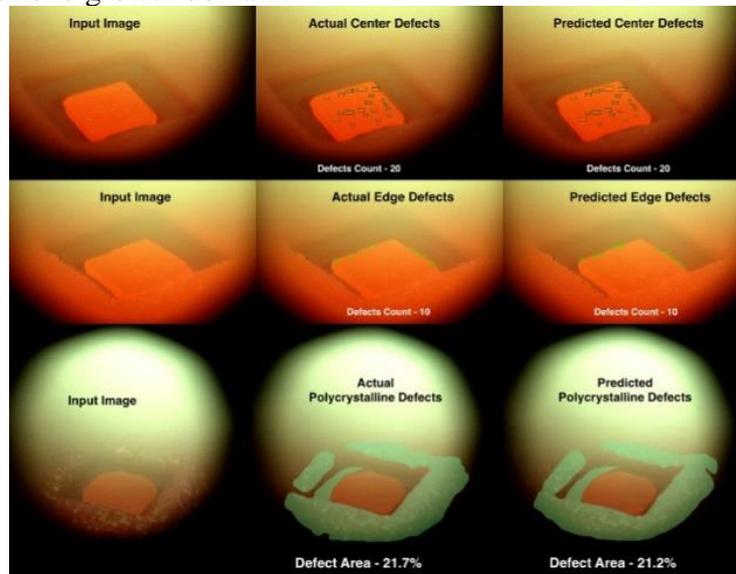

Figure 3. Visual depiction of segmentation and object detection model results on unseen test data; Input Image, Actual/Labeled Defect Image, Predicted Defect Image from left to right;

Step 1: Developing the baseline model - This step builds on the research elaborated on as part of Data Labeling Experiment Module, SP3. and Figure 2 to create an effective baseline model for use in the model assisted labeling (MAL) process. This enables the labelers to correct the model-assisted labels overlayed on image batches instead of having to start annotating each image from scratch. The labeled image batches are used to train the baseline model iteratively until a threshold accuracy is reached as detailed in the previous module.

An essential aspect of MAL baseline model development is the implementation of selective augmentation learning (SAL). During the evaluation phase, images with low segmentation and object detection accuracies undergo selective augmentation using predefined data augmentation techniques [30]. As part of our transformation suite, we integrated a variety of data augmentation methods based on linear transformations like rotation, resizing etc., and nonlinear transformations like noise addition and JPEG compression. These transformations also help simulate unpredictable variations occurring in images due to a host of reasons without having to capture those scenarios as real images during the growth process. For instance, composite rotation and shear based transformations at a smaller scale help perfectly simulate camera vibrations during image capture process. As another example, transformations like emboss, sharpen, blur and Gaussian noise addition help simulate distortions in the image due to stochastic optical hardware problems. Augmented image datasets are then used to re-train the baseline model iteratively. The re-training loop continues for 5 iterations, with low performing images indicating potential data integrity issues. A re-labeling report identifies images for correction by the labeling team. This iterative process repeats until a baseline model surpassing desired accuracy benchmarks (>80%) is achieved. The finalized baseline model is then employed in the MAL process for subsequent image batches, advancing to the next step for the final segmentation and object detection model development.

Step 2: Developing the final model - The second step, as depicted in Figure 4, employs the baseline model developed in Step 1 to pre-annotate all subsequent batches of images for use in development of the final segmentation and object detection models. The labeling team corrects the pre-annotated labels in line with steps outlined in previous section. The entire loop between the image labeling, selective augmentations, re-training and re-labeling through custom reports is repeated until a final model with desired threshold accuracy benchmarks (>95%) is obtained. This final model is then saved to the model database for use in production. Figure 3 graphically depicts results of a successful segmentation and object detection model implementations developed using our pipeline.

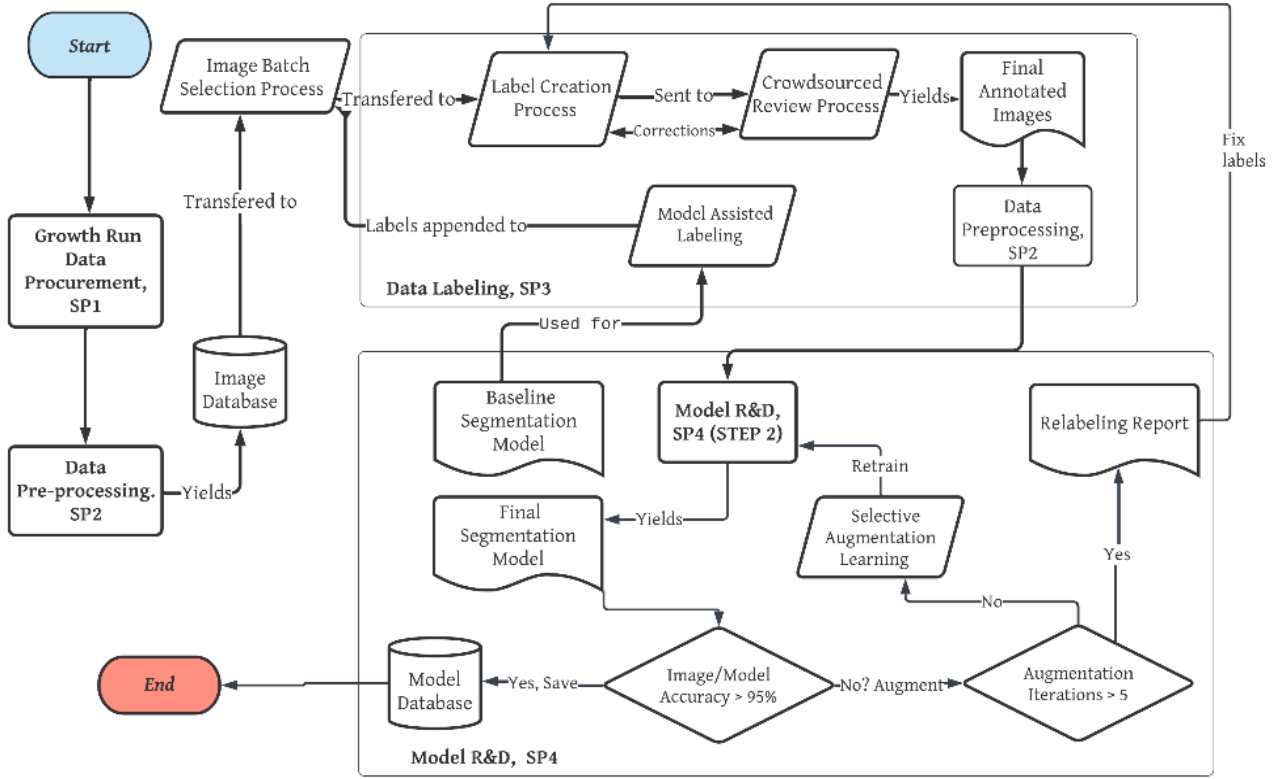

Figure 4. Flow for Step 2 of Model Research and Development Module

**Model Results-Post Analytics**

The model post-analytics module employs the infrastructure and insights gained from SP4 to train model versions based on key hyper-parametric factors of interest like Input image resolution, and Dataset size variations. Overall, we trained 18 different models for the broader semantic segmentation and object detection objective with the above-mentioned hyper-parameter variations. The results of our experiments are detailed in Table I and further elaborated over in subsequent sub-sections.

TABLE I. mAp and IoU based defect detection accuracy on unseen test data

| Model Architecture | | DeeplabV3+ | | YoloV3 |
|---|---|---|---|---|
| Dataset size | Resolution image size | Center-Defects (mAP) | Polycrystalline-Defects (mIoU) | Edge-Defects (mAP) |
| 3000 | 256 × 256 | 60.2 | 87.44 | 70.47 |
| 1500 | 256 × 256 | 58.89 | 87.40 | 69.53 |
| 600 | 256 × 256 | 57.03 | 85.18 | 67.28 |
| 3000 | 512× 512 | **93.35** | **92.83** | **91.98** |
| 1500 | 512× 512 | 90.63 | 88.12 | 90.82 |
| 600 | 512× 512 | 88.54 | 86.18 | 86.02 |

**Input Image Resolution Variations.** In this section, we discuss results of our experiments on understanding the effect of variations in input dataset resolution over the validation accuracy on unseen image data. We employed 2 input image resolutions for this category of experiments - 256x256 and 512x512. The results of these experiments are detailed in Table I.

We observed the best accuracy measures when higher resolutions are used, which can be attributed to the richness of pixel-level features available for higher precision in learnt representations for DOI contours. Table I indicates a performance gain with increase in dataset size when using higher-resolution images. Lower-resolution images, however, experience only marginal improvement in performance with an increase in dataset size.

**Dataset Size Variations.** In this section, we discuss the results of our experiments on understanding the effect of variations in dataset size over the validation accuracy on unseen image data. We employed 3 data augmentation rates for this category of experiments – 600 (2x), 1500 (5x) and 3000 (10x). The results of these experiments are detailed in Table I.

The general trend observed across all model architectures and resolution of images used is that an increase in dataset size contributes to an increase in accuracy. We intend to experiment on this further as part of future research with bigger dataset sizes and augmentation rates.

**Consolidated Results**. Table I details the independent DOI-wise and overall performance breakdown of the models developed across all the hyperparameter variations mentioned in the introduction to this section. Across all the experiments, we obtained maximum DOI-specific accuracies with a dataset size of 3000 and input image resolution of $512 \times 512$ of 93.35% for center defects, 92.83% for polycrystalline defects, and 91.98% for edge defects when tested over an unseen dataset. The best mean accuracy of the defect detection objective was 92.72%. Additionally, our results on the best-performing model show that our pipeline generalizes well on unseen test data even in low dataset volume environments and that an increase in image resolution and dataset size correlates to an incremental gain in model performance. Table I provide a comprehensive overview of the accuracy benchmarks established and of the insights gained from our model development experiments.

**Outlook and future research**

We described in this paper, how we successfully developed and evaluated a novel defect detection pipeline from diamond growth data based on deep learning over a low-volume high-complexity dataset environment constituting use of both object detection and image segmentation algorithms. This approach has achieved state-of-the-art accuracy metrics in classifying and predicting defects. It focuses on semantic image segmentation and object detection to detect the defect label accurately. Additionally, as part of our pipeline, we propose and implement a data development architecture that significantly reduces the time and cost involved and increases the data integrity of the image annotation process. Across all the experiments, we obtained maximum feature of interest-specific accuracy of 93.35% for the Center-Defects, 92.83% for the Poly-crystalline-Defects and 91.98% for the Edge Defects. Additionally, we successfully performed experiments to ascertain key hyper-parameters contributing to optimal precision in our segmentation outputs. For low-volume data experiments like those in crystal synthesis, these results are encouraging and can lead to more widespread implementation of deep learning approaches to crystal process development. As part of future research efforts, we will further test the efficiency of our pipeline by developing more growth run datasets with higher complexity of spatial features in the training data. Additionally, results from the defect extraction models will be used to guide our research on the frame prediction pipeline that predicts future image states of diamond growth from 2 hrs to 16 hrs into the future. The feature and defect extraction models, when applied to the resulting predicted output images from the frame prediction model, will be pivotal in forming automated hypotheses over the effect of reactor parameters like temperature and pressure on defects and feature characteristics. These aspects will be tackled as part of future research to be conducted this year.